\pdfoutput=1

\documentclass[11pt]{article}

\usepackage[]{acl}

\usepackage{times}
\usepackage{latexsym}
\usepackage{adjustbox}
\usepackage{multirow}
\usepackage{amsmath}
\usepackage{graphicx}
\usepackage{subcaption}
\usepackage[dvipsnames]{xcolor}
\usepackage{todonotes}
\usepackage[inline]{enumitem}
\usepackage{paralist}
\usepackage{enumitem}
\usepackage{hyperref}
\usepackage[T1]{fontenc}

\usepackage[utf8]{inputenc}

\usepackage{microtype}

\usepackage{inconsolata}

\title{LASER: An LLM-based ASR Scoring and Evaluation Rubric}

\author{Amruta Parulekar $\quad$ Preethi Jyothi  
\\ 
Indian Institute of Technology Bombay, Mumbai, India \\ 
\small{ \texttt{amrutaparulekar.iitb@gmail.com, pjyothi@cse.iitb.ac.in}} 
} 

\begin{document}
\maketitle
\begin{abstract}

Standard ASR evaluation metrics like Word Error Rate (WER) tend to unfairly penalize morphological and syntactic nuances that do not significantly alter sentence semantics. We introduce an LLM-based scoring rubric LASER that leverages state-of-the-art LLMs' in-context learning abilities to learn from prompts with detailed examples. Hindi LASER scores using Gemini 2.5 Pro achieved a very high correlation score of 94\% with human annotations. Hindi examples in the prompt were also effective in analyzing errors in other Indian languages such as Marathi, Kannada and Malayalam. We also demonstrate how a smaller LLM like Llama 3 can be finetuned on word-pair examples derived from reference and ASR predictions to predict penalty types with close to 89\% accuracy.

\end{abstract}
\begin{table*}[t]
\setlength{\tabcolsep}{6pt}
\centering

\begin{small}
\begin{tabular}{l|c|c}

\hline
\multirow{1}{*}{\textbf{Error type}} &
  
\multicolumn{1}{|c|}{\multirow{1}{*}{\textbf{Example variations}}} &
  \multicolumn{1}{c}{\multirow{1}{*}{\textbf{Penalty}}}  
 
 

   \\ \hline

\multirow{1}{*}{Numerical Phrases} &

   \multirow{1}{*}{"1300" vs "Terah sau" or "Ek hajar teen sau"} &
  \multicolumn{1}{|c}{\multirow{1}{*}{No penalty} }
\\ 
 
 \hline
 \multirow{1}{*}{Abbreviations} &

   \multirow{1}{*}{"ATM" vs "Ay Ti Em" vs "Ay tee yum"} &
  \multicolumn{1}{|c}{\multirow{1}{*}{No penalty} }
\\ 
 
 \hline
 \multirow{1}{*}{Compound Words} &

   \multirow{1}{*}{"bhajan sangraha" vs "bhajansangraha"} &
  \multicolumn{1}{|c}{\multirow{1}{*}{No penalty} }
\\ 
 
 \hline
   \multirow{1}{*}{Transliterations (Native spellings)} &

   \multirow{1}{*}{"ayskreem" vs "aaiskrim" or "skul" vs "skool"} &
  \multicolumn{1}{|c}{\multirow{1}{*}{No penalty} }
\\ 
 
 \hline
   \multirow{1}{*}{Actual transliterations} &

   \multirow{1}{*}{"ice cream" vs "ayskrim" or "aaiskrim"} &
  \multicolumn{1}{|c}{\multirow{1}{*}{No penalty} }
\\ 
 
 \hline
   \multirow{1}{*}{Acceptable alternate spellings} &

   \multirow{1}{*}{"sundar with a bindu" vs "sundar with a half na"} &
  \multicolumn{1}{|c}{\multirow{1}{*}{No penalty} }
\\ 
 
 \hline
 \multirow{1}{*}{Proper nouns} &

   \multirow{1}{*}{"Priya" vs "Pria" vs "Preeya" vs "Preya"} &
  \multicolumn{1}{|c}{\multirow{1}{*}{No penalty} }
\\ \hline
\multirow{1}{*}{Slang and Colloquial terms} &

   \multirow{1}{*}{"Yaha" vs "Ye" or "vaha" vs "vo" or "par" vs "pe"} &
  \multicolumn{1}{|c}{\multirow{1}{*}{No penalty} }
\\ 
 
 \hline
 \multirow{1}{*}{Small (single character) spelling errors} &

   \multirow{1}{*}{"ladki" vs "ladkee" or "bahut" vs "bahoot"} &
  \multicolumn{1}{|c}{\multirow{1}{*}{Minor penalty} }
\\ 
 
 \hline
  \multirow{1}{*}{Small grammatical errors (gender/tense/number)} &

   \multirow{1}{*}{"hain" vs "hai" or "uska" vs "uski" vs "usko"} &
  \multicolumn{1}{|c}{\multirow{1}{*}{Minor penalty} }
\\ 
 
 \hline
   \multirow{1}{*}{Spelling errors that alter meaning} &

   \multirow{1}{*}{"kumar" vs "kamar" or "saman" vs "samanya"} &
  \multicolumn{1}{|c}{\multirow{1}{*}{Major penalty} }
\\ 
 
 \hline
     \multirow{1}{*}{Incorrect word substitutions} &

   \multirow{1}{*}{"sundar" vs "bhadda" or "mota" vs "chhota"} &
  \multicolumn{1}{|c}{\multirow{1}{*}{Major penalty} }
\\ 
 
 \hline
   \multirow{1}{*}{Significant omissions or additions} &

   \multirow{1}{*}{"--" vs "sundar" or "mota" vs "--"} &
  \multicolumn{1}{|c}{\multirow{1}{*}{Major penalty} }
\\ 
 
 \hline
   \multirow{1}{*}{Reordering of words that changes meaning} &

   \multirow{1}{*}{"bahut accha khana" vs "bahut khana accha"} &
  \multicolumn{1}{|c}{\multirow{1}{*}{Major penalty} }
\\ 
 
 \hline

\end{tabular}%

\end{small}

\caption{Types of ASR errors and their penalties.}
\vspace{-10pt}
\label{tableerr}
\end{table*}

\section{Introduction and Related Work}

Automatic Speech Recognition (ASR) is used in a variety of applications ranging from voice assistants~\cite{schwarz2023personalizedpredictiveasrlatency} to accessibility aids~\cite{50459}. This makes it increasingly important to design accurate evaluation metrics for ASR systems. The most widely used ASR evaluation metric is word error rate (WER) (or character error rate, i.e., CER, for languages that do not have well-defined word boundaries). WER/CER are edit distance-based metrics that compute the minimum number of substitutions, insertions and deletions needed to transform an ASR prediction to its corresponding reference transcription. Lexically-sensitive metrics that are based on exact matches like WER penalize a prediction even if the ASR error is very minor in nature. This limitation of WER gets further amplified for Indian languages.

\vspace{3pt}

Several characteristics of Indian languages render WER a sub-optimal ASR evaluation metric:
\begin{enumerate*} 
\item Many Indian languages are morphologically rich with words having many inflectional variants~\cite{vikrammorphology}, words containing gender/tense/number markers~\cite{Pitale_Sarma_2013}, words being agglutinative in nature~\cite{Krishnamurti_2003}, etc. ASR predictions might contain minor errors in terms of these morphological inflections which get treated as major errors by WER. 
\item Compound words are common across many Indian languages \cite{Kulkarni2012SemanticPO}. There are multiple accepted forms of writing the same word (e.g., \textit{paas wala} vs. \textit{paaswala}). WER treats one form as an error if the reference contains the other.
\item Many Indian languages contain English loan words that do not have standardized native script spellings. (E.g., ice cream in Devanagari could be written as \textit{ayskrim} or \textit{aaiskreem}). Although such variants should be treated the same, WER penalizes any variant differing from the reference. 
\end{enumerate*}
There are other error types common to English and Indian languages like colloquialisms (dunno vs. don't know), abbreviations (brb vs. be right back), numerical phrases (10 vs. ten), etc. that should also ideally incur no penalty during evaluation. 
\vspace{3pt}

Semantic metrics like BERTScore \cite{zhang2020bertscoreevaluatingtextgeneration} or SemDist \cite{kim2021semanticdistancenewmetric} are based on embeddings and do not always fare well on alternate spellings. Phoneme Error Rate (PER) \cite{Yolchuyeva_2019} and CER accommodate alternate phoneme/character-based spellings but treats all errors equally, regardless of semantic impact. Thus, there is a need for a nuanced evaluation metric that heavily penalizes semantically significant errors, lightly penalizes minor ones, and ignores acceptable variations. Large language models (LLMs), with their strong in-context learning abilities, can be leveraged for this purpose.

\vspace{3pt}

In this work, we propose a novel LLM-based scoring and evaluation rubric for ASR (LASER). LASER avoids penalizing colloquial spelling variations, compound words, alternate transliteration spellings, and variant representations of numbers and abbreviations. It applies minor penalties to spelling or grammatical errors that preserve sentence meaning, and major penalties to word insertions, omissions, and meaning-altering errors. This was achieved via a carefully curated prompt to state-of-the-art LLMs and the LLM scores were compared with scores from humans who were given the same instructions. LASER correlates very well with human scores, unlike WER. Cross-lingual tests assessed whether multilingual LLMs could transfer this knowledge from high- to low-resource languages; interestingly, a prompt with Hindi examples transfers well to other Indian languages like Marathi, Kannada and Malayalam. 
Finally, we tested whether an open-source LLM could be trained on a word-pair dataset curated using LLM outputs to our prompt, to make our metric and penalization strategy publicly available.\footnote{Our scripts and checkpoints to use LASER for Hindi available at \href{https://github.com/Amparulekar/LASER-metric}{https://github.com/Amparulekar/LASER-metric}} LASER is an open-source, LLM-based fine-grained scoring metric for ASR, which attains high agreement with human evaluations.
\vspace{3pt}

In recent work, \citealp{10447177} designed LATTEScore, an LLM-based metric that assesses meaning preservation in ASR transcripts of impaired speech through classification of sentences based on whether their meaning is preserved. \citealp{phukon2025} is concurrent work aligned with our primary objective; it combines natural language inference scores with semantic similarity and phonetic similarity to evaluate  logical similarity between the prediction and ground truth. However, it focuses on correctability of dysarthric English speech while we focus on a metric that accounts for linguistic nuances of different languages. 

\vspace{-6pt}
\section{Methodology}
\label{method}
\vspace{-3pt}
\paragraph{Metric development.}
To develop LASER, we first analyzed error types penalized by standard ASR metrics and assigned revised penalties: lower for minor grammatical errors and higher for semantic errors. Semantically equivalent variations, like compound words and transliterations, should incur no penalty. Table \ref{tableerr} lists the no-penalty, minor-penalty, and major-penalty error types that we identified. Major and minor errors were assigned penalties of 1 and 0.5 points, respectively. Non-penalizable errors incurred no penalty. A sentence-level score is then defined as $1 - \frac{\text{Total penalty}}{\text{Number of reference words}}$.

\paragraph{Prompt development.}
Our main LLM prompt (Appendix \ref{a}) is in English (Latin script) with examples in Hindi (Devanagari script). 
Our prompt has three main components:
\begin{compactenum}[(a)]    
\item Detailed instructions: The LLM is instructed to tokenize sentences by words, align ground-truth labels with predictions forming word pairs, and identify mismatches. The LLM is further instructed to classify the mismatches by error type via detailed examples and assign major, minor or no penalty, and finally, compute the sentence-level LASER score by adding up the penalties.
\item Detailed examples: We provided an example for every error type (shown in Table~\ref{tableerr}).
\item Promote step-by-step reasoning: We promoted chain-of-thought reasoning by asking the LLM to return its response in the format:  
\textit{(Word count of original sentence; list of non-penalizable errors; list of major penalizable errors; list of minor penalizable errors; total penalty; score)}. This format ensured that the LLM applied the metric consistently.
\end{compactenum}

\paragraph{Dataset creation for LLM Finetuning.}
Since LLM API calls are expensive, we investigated whether we could finetune a smaller LLM (e.g., Llama 3.1) with aligned word pairs (derived automatically by aligning the ASR output and the reference) to predict whether the word-pair incurs a major, minor or no penalty. Word pairs for training are obtained via human-annotated transcripts. Section ~\ref{classi} provides more details of this experiment.



\begin{figure}[t] 
  \centering

\includegraphics[width=0.95\linewidth]{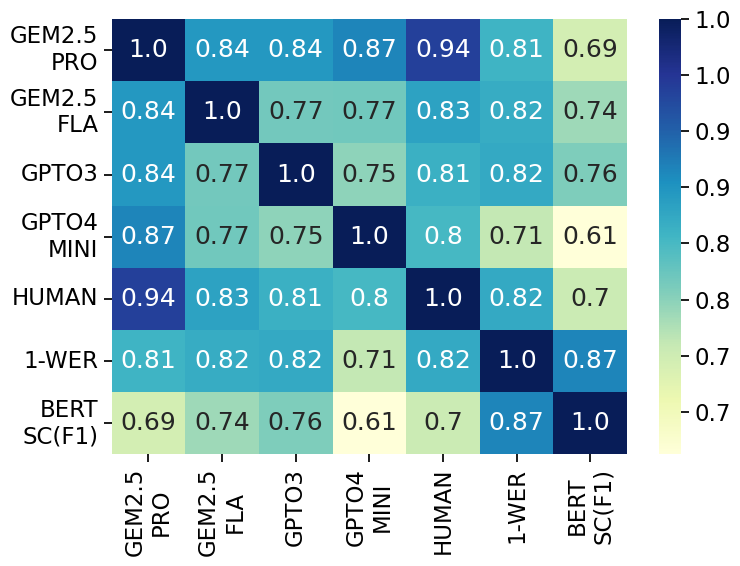}

  \caption{Correlation heatmap for different LLM scores using the Hindi prompt, Human scores, WER and BERTScore(F1) on Hindi data.}
  \label{fighin}

  \end{figure}

\section{Experimental Setup}

\label{setup}
\textbf{Dataset.}
We used a subset of the IndicVoices test set \cite{javed2024indicvoicesbuildinginclusivemultilingual}, a multilingual, multi-speaker collection of natural, spontaneous speech. We used the multilingual SeamlessM4T \cite{communication2023seamlessm4tmassivelymultilingual} model to generate ASR predictions. Sentence pairs with no transcription mismatch (i.e., 0 WER) were removed. We focused on two Indo-Aryan (Hindi, Marathi) and two Dravidian (Malayalam, Kannada) languages. Our final datasets had 172 Hindi, 154 Marathi, 229 Malayalam, and 216 Kannada sentence pairs. 
\vspace{6pt}

\noindent\textbf{LLMs.} For prompt-tuning, we chose LLMs known for their strong reasoning capabilities. From the Gemini \cite{geminiteam2024gemini15unlockingmultimodal} and GPT \cite{openai2024gpt4technicalreport} families, Gemini 2.5 Pro and GPTo3 are advanced reasoning models, while Gemini 2.5 Flash and GPTo4mini offer speed and cost-efficiency.\footnote{Deepseek R1 failed to produce scores using the prompt.} For finetuning of the word-pair classification task, we chose the open-source Llama 3 8B. \cite{grattafiori2024llama3herdmodels}.
\newpage

\noindent\textbf{Score evaluation.}
To evaluate LLM outputs, we assigned the same task to humans using identical instructions, examples, and a worked-out sample. The humans were paid Rs. 24  per sentence to list sentence-wise major, minor and non-penalizable errors and their counts; more details are in Appendix \ref{b}. These penalty counts were used to calculate our LASER scores. We computed the Pearson's correlation coefficients between human, LASER, and standard metric scores. A higher human-LASER score correlation compared to human-WER correlation would indicate that the LASER scores are more accurate.

\section{Experiments and Results}

\subsection{Correlation analysis}

 \begin{figure}[t]
\centering
\vspace{2pt}
\begin{subfigure}{.235\textwidth}
    \centering
    \includegraphics[width=\linewidth]{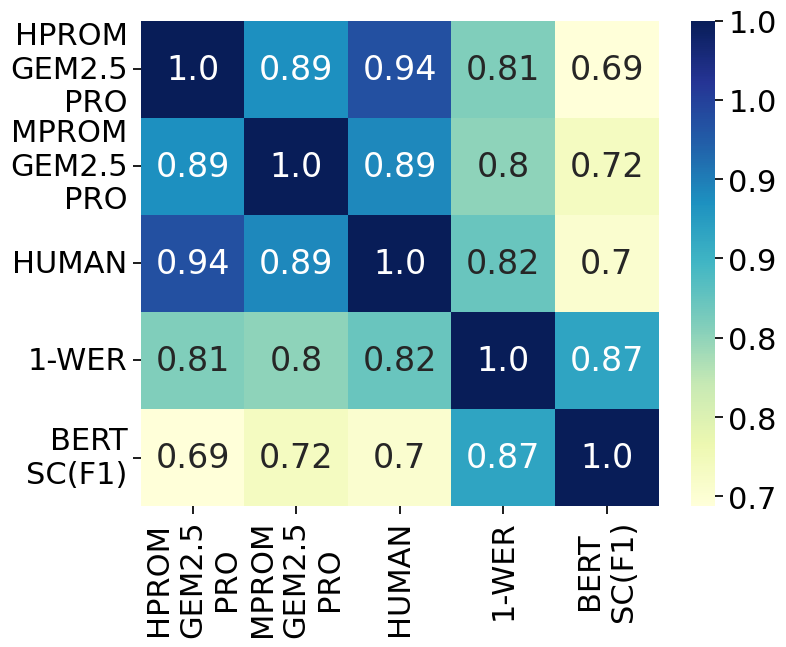}  
  
    \caption{Hindi data}
    \label{SUBFIGURE LABEL A}
\end{subfigure}
\begin{subfigure}{.235\textwidth}
    \centering
    \includegraphics[width=\linewidth]{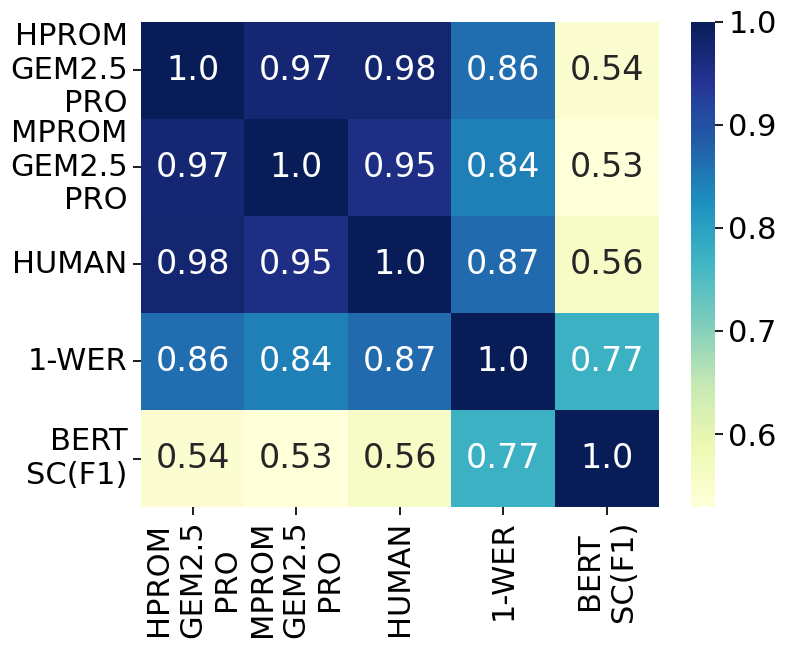} 
    
    \caption{Marathi data}
    \label{SUBFIGURE LABEL B}
\end{subfigure}

\caption{Correlation heatmap for Human scores, WER, BERTScore(F1) and Gemini 2.5 Pro scores using the Hindi \textit{(HPROM)} and the Marathi \textit{(MPROM)} prompts for both the Hindi and the Marathi data.}
\vspace{-6pt}
\label{FIGURE LABEL A}
\end{figure}
  
    \begin{figure*}[t]
\centering
\begin{subfigure}{.325\textwidth}
    \centering
    \includegraphics[width=\linewidth]{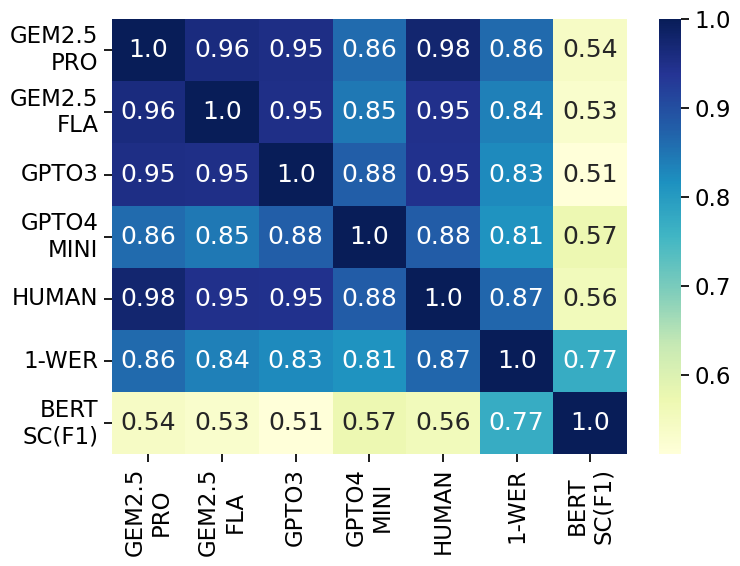}  
    \vspace{-17pt}
    \caption{Marathi}
    \label{SUBFIGURE LABEL 1}
\end{subfigure}
\begin{subfigure}{.325\textwidth}
    \centering
    \includegraphics[width=\linewidth]{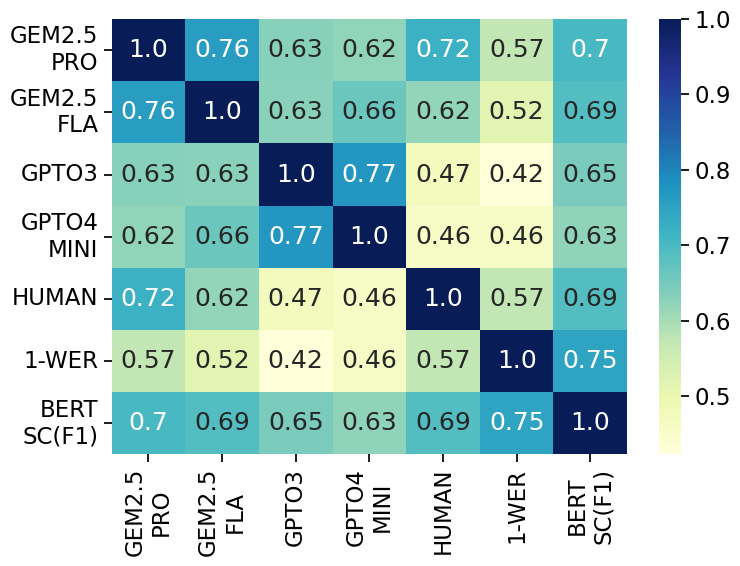}  
    \vspace{-17pt}
    \caption{Malayalam}
    \label{SUBFIGURE LABEL 2}
\end{subfigure}
\begin{subfigure}{.325\textwidth}
    \centering
    \includegraphics[width=\linewidth]{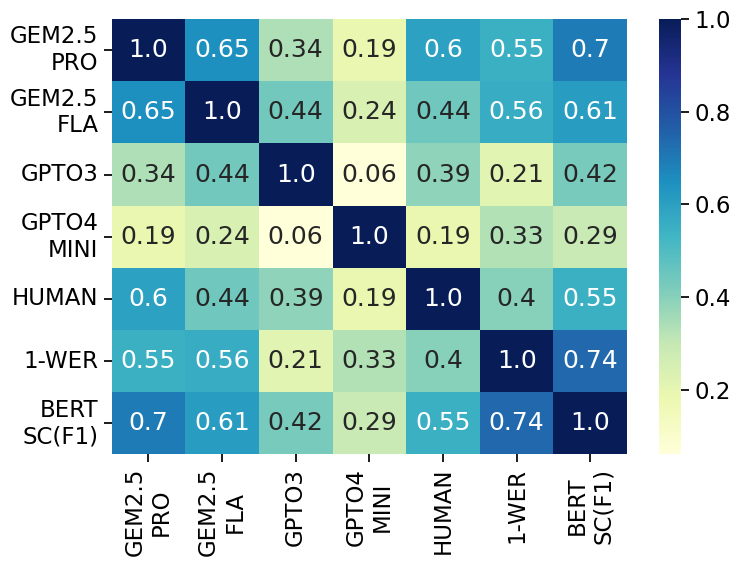}  
    \vspace{-17pt}
    \caption{Kannada}
    \label{SUBFIGURE LABEL 3}
\end{subfigure}
\vspace{-7pt}
\caption{Correlation heatmaps for different LLM scores using the Hindi prompt, Human scores, WER and BERTScore(F1) for  154 Marathi, 229 Malayalam and 216 Kannada sentences.}
\vspace{-6pt}
\label{FIGURE LABEL}
\end{figure*}

\begin{figure*}[t] 
\centering
\includegraphics[clip,width=\linewidth,trim=3.5cm 11.5cm 4cm 2cm]{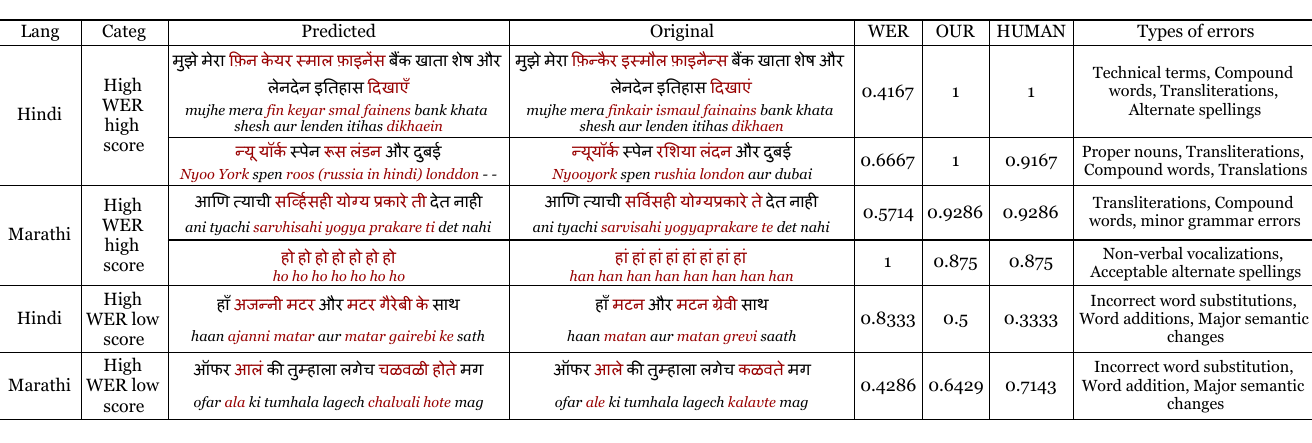}
\vspace{-17pt}
  \caption{Qualitative analysis of high WER samples having high and low LASER scores. \textcolor{BrickRed}{Red text} indicates mismatch between the original and predicted transcriptions.}
  \vspace{-5pt}
  \label{figtab}

  \end{figure*}


Figure \ref{fighin} depicts the correlation heatmap of LLM-based LASER scores, human scores and standard metrics (WER, BERTScore) for 172 Hindi sentences. Gemini 2.5 Pro outperformed every other LLM, exhibiting the highest correlation with human scores and significantly surpassing WER correlation scores. Additionally, Gemini 2.5 Pro consistently recalled the initial scoring instructions, showed clear reasoning, and formatted results correctly after each batch. Notably, it was also able to infer appropriate penalties for error types not included in the prompt, viz. sandhi (phonetic transformation at word boundaries during word fusion) \cite{dave2020neuralcompoundwordsandhigeneration} and synonyms.
\vspace{6pt}

IndicVoices is primarily noisy and contains a majority of spontaneous/conversational speech. It is also rich in dialectal diversity with speech covering 145 Indian districts \& 22 languages. Our experiments demonstrate that LASER performs well on noisy \& dialectal speech.
%


\newpage
\vspace{-3pt}
\subsection{Cross-lingual transfer}

We developed a new prompt having English instructions with Marathi examples similar to the Hindi prompt examples. To compare the efficacy of cross-lingual transfer between higher- (Hindi) and lower resource (Marathi) languages of the same language family (Indo-Aryan), we used the Marathi and the Hindi prompts on both the Hindi and the Marathi sentences. Only Gemini 2.5 Pro was used, as it was the best-performing LLM for both languages. Figures \ref{SUBFIGURE LABEL A} and \ref{SUBFIGURE LABEL B} show the correlations using both prompts on the Hindi and the Marathi sentences respectively. Although both prompts gave high human score correlations, the Hindi prompt performed better for both languages, likely due to the LLM's familiarity with Hindi.

To evaluate cross-lingual inference, i.e., whether the LLM could infer mismatch types in a new language given examples of a different language, the prompt with Hindi examples was used to obtain scores for Marathi, Malayalam and Kannada. Malayalam and Kannada are lower-resource Dravidian languages that are syntactically and morphologically very different from Hindi. The scoring method and the LLMs were the same as those for Hindi. Figures \ref{SUBFIGURE LABEL 1}, \ref{SUBFIGURE LABEL 2} and \ref{SUBFIGURE LABEL 3} depict the correlation heatmaps of Hindi-prompted LLM scores, standard metrics and human scores for the three languages. The Hindi prompt transferred effectively and we observe trends similar to Hindi. Gemini 2.5 Pro was the best performing LLM, yielding the highest correlation with human scores. This indicates that LLMs like Gemini 2.5 Pro are able to adapt grammar rules from one language to another (even from a different language family). The correlation scores for Marathi were higher than those for Hindi; this could be due to the shorter average sentence length (20.86 words for Marathi vs. 27.34 words for Hindi) which may have reduced processing complexity. The correlation scores for Malayalam and Kannada were lower than those for Hindi and Marathi, potentially due to nuances of Dravidian languages that the Hindi prompt was unable to address. Notably, even on Dravidian languages, Gemini 2.5 Pro was able to consider target-language nuances beyond those explicitly included in the Hindi prompt. This experiment demonstrates that our carefully designed prompt can be scaled to multiple languages. We also ran an experiment on English, details of which are in Appendix~\ref{d}.

\newpage

\subsection{Qualitative Analysis}

We qualitatively analyzed high WER samples (greater than 0.35) that had high and low LASER scores. On instances with high WERs, we checked whether our metric does indeed correct the unfair penalization of semantically identical but syntactically different mismatches between references and ASR predictions. Figure \ref{figtab} shows how high WER and high LASER score samples contained a high percentage of non-penalizable errors of different types (while retaining the word-pair meanings); in contrast, low LASER score samples had significant semantic word-pair mismatches. Moreover, human scores are consistent with the LASER scores. This validates the necessity and utility of our metric.
\vspace{-5pt}

\subsection{Finetuning for word-pair classification}

\label{classi}
To develop a more efficient way to use our metric, we performed low-rank adaptation (LoRA) finetuning of the Llama3-8B model on a word-pair classification objective for the classes - "No mismatch", "Non-penalizable error", "Major penalty" and "Minor penalty". The LoRA model contains 3.4M trainable parameters. We used 950 word-pairs to finetune this model. 

Evaluation was done in two ways: 1) test-train split of the train set (to evaluate classification accuracy), 2) holding out 17 out of 172 sentences as a test set prior to train set creation (to evaluate scoring efficacy). A Hindi word-pair classification dataset was curated after manual corrections to Gemini 2.5 Pro outputs and adding a random set of no-mismatch pairs. \\

\begin{table}[htbp]
\setlength{\tabcolsep}{4pt}
\centering
\vspace{-15pt}
\begin{small}
\begin{tabular}{l|c|c|c|c}

\hline
\multirow{1}{*}{\textbf{Class}} &
  \multicolumn{1}{|c|}{\multirow{1}{*}{\textbf{\#train+val}}} &
\multicolumn{1}{|c|}{\multirow{1}{*}{\textbf{\#test}}} &
  \multicolumn{1}{|c|}{\multirow{1}{*}{\textbf{\#Correct}}}  
  &
  \multicolumn{1}{c}{\multirow{1}{*}{\textbf{Accuracy}}}
 
 

   \\ \hline

\multirow{1}{*}{0 (Identical)} &
\multirow{1}{*}{310} &
   \multirow{1}{*}{34} &
 \multirow{1}{*}{32} & \multirow{1}{*}{94.12\%} 
\\ 
 \multirow{1}{*}{1 (Non-Pen)} &
\multirow{1}{*}{312} &
   \multirow{1}{*}{35} &
 \multirow{1}{*}{31} & \multirow{1}{*}{88.57\%} 
\\ 
 \multirow{1}{*}{2 (Minor)} &
\multirow{1}{*}{77} &
   \multirow{1}{*}{9} &
 \multirow{1}{*}{6} & \multirow{1}{*}{66.67\%} 
\\ 
\multirow{1}{*}{3 (Major)} &
\multirow{1}{*}{251} &
   \multirow{1}{*}{28} &
 \multirow{1}{*}{25} & \multirow{1}{*}{89.29\%} 
\\ 
\hline
\multirow{1}{*}{All} &
 \multirow{1}{*}{950} &
   \multirow{1}{*}{106} &
 \multirow{1}{*}{94} & \multirow{1}{*}{88.69\%} 
\\ 
 
 \hline

\end{tabular}%

\end{small}

\caption{ Class-wise accuracies on finetuning Llama3-8.}
\vspace{-7pt}
\label{tablecals}
\end{table}

\noindent\textbf{Test-train split.} 
 Table \ref{tablecals} depicts the model's test set accuracies, achieving 88.69\% across all classes. We observe that that the minor-penalty errors are the toughest for Llama to identify, as they are more infrequent compared to the other error categories.
\begin{figure}[htbp] 
\centering

  \includegraphics[width=0.6\linewidth]{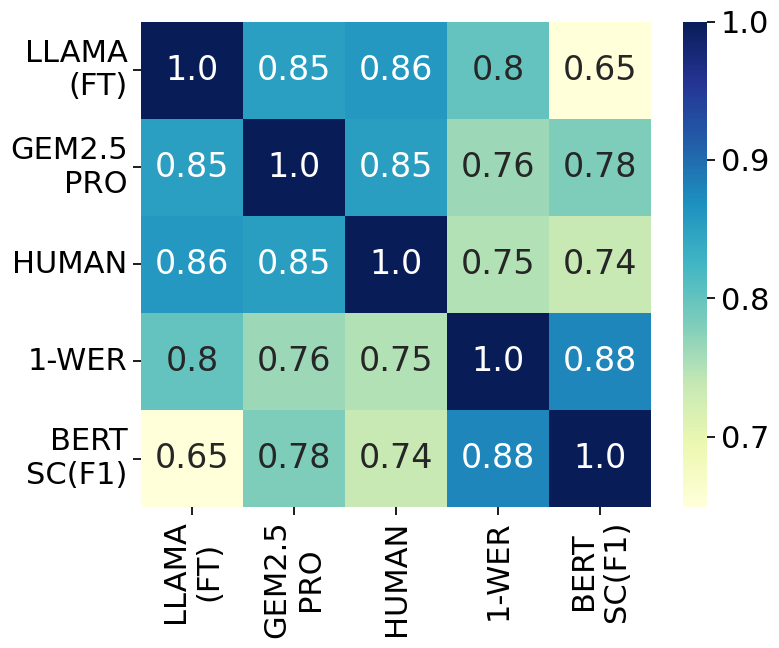}
  \vspace{-5pt}
  \caption{Correlation heatmap for finetuned Llama3, Gemini 2.5 Pro (Hindi prompt), Human, WER and BERTScore(F1) for 17 held-out Hindi sentence pairs.}
  \vspace{-12pt}
  \label{figlam}
  \end{figure}
\vspace{3pt}  

\noindent\textbf{Held-out sentences.}  The 17 held-out sentence pairs were aligned into corresponding word pairs using a custom greedy alignment script. These aligned pairs were converted into a test set to evaluate the LLM. Llama predictions were used to obtain a total penalties and corresponding scores for the 17 sentences. Figure \ref{figlam} depicts the correlation heatmap of Llama3 scores, Gemini-2.5-Pro scores, human scores and standard metrics for the 17 held out sentences. Llama performs even better than Gemini-2.5-Pro and is more aligned with human scores, potentially due to the manual corrections of the Gemini-2.5-Pro outputs prior to training.
\vspace{-4pt}
\section{Conclusion}
\vspace{-4pt}
In this work, we develop LASER, a fine-grained LLM-driven ASR metric that considers semantic, linguistic and morphological nuances and does not unfairly penalize predicted transcriptions. We use a carefully curated prompt with detailed descriptions of error types in Hindi. We tested the prompt on multiple LLMs and compared the results with human evaluations. We observe that LLMs like Gemini-2.5-Pro are very well-correlated with human annotations unlike standard measures like WER. We are also able to use the prompt with Hindi examples to effectively transfer knowledge to transcriptions in other languages from the same as well as different language families (Marathi, Kannada, Malayalam). Finally, we show the feasibility of a more efficient evaluation setup by finetuning Llama-3 to learn our penalty rules using a small amount of hand-annotated data. 
\vspace{-4pt}
\section*{Limitations}
\vspace{-4pt}
LLMs tend to process ambiguities differently on different runs. For instance, a slang spelling might be considered a spelling error and penalized incorrectly. It was observed that these differences were higher in case of lower-resource languages. Although this occurs in a small number of cases and the variation is small, there is a need to develop a standardized technique that will ensure the same score on all runs. Finetuning Llama with Gemini predictions  addresses the LLM inference inconsistency issue, as the weights can be fixed at inference time to obtain consistent outputs. 

Our prompt-based technique has higher latency compared to other metrics. We improved the efficiency of LASER through Low-rank adaptation (LoRA) finetuning of Llama, but reducing latency further can be a direction of future research.

\bibliography{anthology}

\appendix

\section{LLM Prompt (Hindi)}
\label{a}
Italicized text in the prompt below was in Devanagari script in the original instructions.
 \vspace{6pt}
 
\noindent \textbf{YOUR TASK: } 
 \vspace{6pt}
\\Here's how we can address these challenges in Indian language ASR evaluation and design a scoring metric system from 0 to 1. Designing the Metric -
 \vspace{3pt}
\\A) Define Non-Penalizable Errors: 
 \vspace{3pt}
\begin{compactenum}
    \item Numbers: Accept different spellings for numbers (e.g., "1300", "thirteen hundred", "\textit{Terah sau}").
    \item Abbreviations: Accept variants in spelling abbreviations (e.g., ATM spelled as
"\textit{aytiem}" or "\textit{ayteeyam}").
\item Compound Words: Accept variations in joining or separating compound words (e.g.,
"\textit{bhajan sangraha}" vs. "\textit{bhajansangraha}" or "\textit{paas wala}" vs. "\textit{paaswala}").
\item Native Spellings of Transliterated Words: Accept variants in spelling transliterations (e.g.,"\textit{aaiskrim}" vs. "\textit{aaiskreem}" vs. "\textit{ayskrim}")
\item Transliterated words: Accept latin script spellings of transliterated words
(e.g.,"\textit{aiskrim}" vs. Ice cream or "\textit{skool}" vs. School)
\item Alternate Spellings: Accept grammatically correct spelling differences (e.g., "\textit{sundar (with a bindu)}" vs.
"\textit{sundar (with half "na"})".
\item Proper Nouns: Allow for minor spelling variations in names and places (e.g., "\textit{priya}"
vs. "\textit{preya}" vs. "\textit{preeya}").
\item Slang and Colloquial Terms: Account for regional variations (e.g., "\textit{yaahaan}" vs. "\textit{yaahaa}" or
"\textit{yaha}" vs. "\textit{ye}").
\end{compactenum}
 \vspace{3pt}
\noindent B) Define Minor Penalizable Errors (0.5 points): 
 \vspace{3pt}
\begin{compactenum}
    \item Small spelling error: Minor penalty for small single character spelling errors that
sound similar (e.g. "\textit{ladkee}" vs. "\textit{ladki}")
\item Small grammatical error: Minor penalty for a small grammatical error that does not
alter meaning (e.g. Gender error or singular plural error like "\textit{uska} " vs. " \textit{uski}" or "\textit{hain}"
vs. "\textit{hai}")
\end{compactenum}
 \newpage
\noindent C) Define Major Penalizable Errors (1 point): 
 \vspace{3pt}
\begin{compactenum}
    \item Incorrect word substitutions (e.g., replacing "\textit{sundar}" with "\textit{bhadda}").
    \item Significant omissions or additions.
    \item Reordering of words that changes the meaning.
    \item Spelling mistakes that change meaning (e.g. “\textit{kumar}” vs. “\textit{kamar}”)
\end{compactenum}
 \vspace{3pt}
D) Matching Strategy: 
 \vspace{3pt}
\begin{compactenum}
    \item Token-Based Matching: Use fuzzy matching to compare each token (word). Assign weights to tokens to prioritize penalizable errors over non-penalizable ones. 
    \item Phonetic Similarity: Leverage phonetic matching algorithms like Soundex or Metaphone to compare pronunciation. 
\end{compactenum}

\vspace{3pt}
\noindent E) Scoring: 
\vspace{3pt}

\begin{compactenum}
    \item $Score = ( 1 - \frac{\text{penalized errors}}{\text{total tokens}}$ ).
    \item Weight errors differently based on severity (e.g., minor spelling variation = 0.5, word substitution = 1.0). 
\end{compactenum}
Once we provide the two sentences, apply the above rules and give a similarity score between 0 and 1.  
\vspace{6pt}
 
\noindent \textbf{EXAMPLE: }  
\vspace{6pt}

\noindent \textbf{Step 1: Tokenization } 
\vspace{3pt}
\\Sentence 1 - predicted:
\\\textit{vaha}, \textit{bhajan}, \textit{sangraha}, \textit{komal}, \textit{paaswala}, \textit{aytiem}, 10, \textit{par}, \textit{taims}, \textit{sundar (with bindu)}, \textit{hain}, \textit{skul}, \textit{se} 
\vspace{3pt}
\\Sentence 2 - original:
\\\textit{vo}, \textit{bhajansangraha}, \textit{ke} , \textit{paas}, \textit{walaa}, A.T.M., \textit{das}, times, \textit{sundar(with half na)}, \textit{hai}, \textit{skool}, \textit{se}
\vspace{3pt}
\\\noindent \textbf{Step 2: Classify Tokens } 
\vspace{3pt}
\\A) Non-penalizable errors:
\vspace{3pt}
\begin{compactenum}
    \item Colloquial variations:
\\\textit{vaha} vs. \textit{vo}: Colloquial difference - NO PENALTY
\item Compound word handling:
\\\textit{bhajan sangraha} vs. \textit{bhajansangraha}, \textit{paaswala} vs. \textit{paas wala}: Acceptable compound word variations - NO PENALTY
\item Abbreviation variations:
\\\textit{aytiem} vs. A.T.M.: Abbreviation handling - NO PENALTY
\item Numerical variation:
\\10 vs. \textit{das}: Equivalent numerical representation - NO PENALTY
\newpage
\item Transliterations:
\\\textit{taims} vs times: Acceptable transliteration difference - NO PENALTY
\item Alternate spellings:
\\isundar (with bindu) vs. \textit{sundar (with half na)}: Regional spelling difference - NO PENALTY
\item Transliteration spelling variations:
\\\textit{skul} vs. \textit{skool}: Acceptable transliteration difference - NO PENALTY
\end{compactenum}
\vspace{3pt}
B) Penalizable errors:
\vspace{3pt}
\begin{compactenum}
    \item \textit{komal} vs. \textit{ke}:   
Wrong substitution of word - Penalty weight = 1.0 (major error) 
\item \textit{par}: 
Addition of word 
- Penalty weight = 1.0 (major error) 
\item \textit{hain} vs. \textit{hai}: 
Small grammatical error that changes singular to plural. 
- Penalty weight = 0.5 (minor error)  
\end{compactenum}
\vspace{3pt}
\noindent C) Exact matches:  
\vspace{3pt}
\begin{compactenum}
    \item \textit{se}: Appears identically in both - No penalty 
\end{compactenum}
\vspace{3pt}
 
\noindent \textbf{Step 3: Scoring } 
\vspace{3pt}
\\The formula is:  
$ \text{Score} = 1 - \frac{\text{Weighted penalized errors}}{\text{Total tokens}} $  
\\Total tokens: 12 (from Sentence 2)   
\\Penalized errors:   
\\\textit{komal} vs. \textit{ke} : 1.0, \textit{par}: 1.0, \textit{hain} vs. \textit{hai}: 0.5 
\\Weighted penalized errors: (1.0 + 1.0 + 0.5 = 2.5)  
\\$ \text{Score} = 1 - \frac{2.5}{12} = 1 - 0.2083 = 0.7917 $  
\\Final Similarity Score: 0.7917 
\vspace{6pt}
\\\textbf{STRUCTURE OF YOUR RESPONSE: }
\vspace{6pt}
\\Number of tokens in original sentence; list of tokens with non-penalizable errors; list of tokens with major penalizable errors; list of tokens with minor penalizable errors; total penalty; score. If I give you predicted and original sentence pairs, can you return ONLY the output I asked for in a single json (number each sentence pair) and not details.

\section{Human instructions}
\label{b}

Human annotators, who had thorough linguistic knowledge of the languages that we worked on, were commissioned to annotate our sentence pairs and obtain total penalties for each sentence. They charged us Rs. 24 per sentence pair, for the 172 Hindi, 154 Marathi, 229 Malayalam and 216 Kannada sentence pairs. Italicized text in the instructions below was in Devanagari script in the original instructions.
\newpage

\noindent \textbf{INSTRUCTIONS:} 
\vspace{6pt}

\noindent The annotator should look at the two sentences side by side (original and predicted) and look for any difference between the two sentences. Each difference is an error. Now of these errors, we have classified them into 3 errors, no penalty, minor penalty and major penalty. The types of errors and their classification is explained in the rules. For instance if one sentence has 1300 and the other has terah sau written, this is a no penalty error. So we need the annotators to make lists for each sentence pair, of the major, minor and no penalty errors (3 lists). Each list must be of the format ‘1300 vs terah sau (numbers), vaha vs vo (colloquial variation)’ and so on i.e. word in first sentence vs word in second sentence and the reason why they are classified in this error type. And finally we also need the counts of the types of errors for each sentence pair.
\vspace{6pt}

\noindent \textbf{RULES:}
\vspace{6pt}
\\A) NO-PENALTY Errors:
\vspace{3pt}
\begin{compactenum}
    \item Numbers: Accept different spellings for numbers (e.g., "1300", "thirteen hundred", "\textit{Terah sau}").
    \item Abbreviations: Accept variants in spelling abbreviations (e.g., ATM spelled as
"\textit{aytiem}" or "\textit{ayteeyam}").
\item Compound Words: Accept variations in joining or separating compound words (e.g.,
"\textit{bhajan sangraha}" vs. "\textit{bhajansangraha}" or "\textit{paas wala}" vs. "\textit{paaswala}").
\item Native Spellings of Transliterated Words: Accept variants in spelling transliterations
(e.g.,"\textit{aaiskrim}" vs. "\textit{aaiskreem}" vs. "\textit{ayskrim}")
\item Transliterated words: Accept latin script spellings of transliterated words
(e.g.,"\textit{aiskrim}" vs. Ice cream or "\textit{skool}" vs. School)
\item Alternate Spellings: Accept grammatically correct spelling differences (e.g., "\textit{sundar (with a bindu)}" vs.
"\textit{sundar (with half "na"})".
\item Proper Nouns: Allow for minor spelling variations in names and places (e.g., "\textit{priya}"
vs. "\textit{preya}" vs. "\textit{preeya}").
\item Slang and Colloquial Terms: Account for regional variations (e.g., "\textit{yaahaan}" vs. "\textit{yaahaa}" or
"\textit{yaha}" vs. "\textit{ye}").
\end{compactenum}
\newpage

\noindent B) MINOR-PENALTY Errors:
\vspace{3pt}
\begin{compactenum}
    \item Small spelling error: Minor penalty for small single character spelling errors that
sound similar (e.g. "\textit{ladkee}" vs. "\textit{ladki}")
\item Small grammatical error: Minor penalty for a small grammatical error that does not
alter meaning (e.g. Gender error or singular plural error like "\textit{uska} " vs. " \textit{uski}" or "\textit{hain}"
vs. "\textit{hai}")
\end{compactenum}
 \vspace{3pt}
 
\noindent C) MAJOR-PENALTY Errors:
 \vspace{3pt}
 
\begin{compactenum}
    \item Incorrect word substitutions (e.g., replacing "\textit{sundar}" with "\textit{bhadda}").
    \item Significant omissions or additions.
    \item Reordering of words that changes the meaning.
    \item Spelling mistakes that change meaning (e.g. “\textit{kumar}” vs. “\textit{kamar}”)
\end{compactenum}
 \vspace{6pt}
\noindent \textbf{EXAMPLE:}
 \vspace{6pt}
\\Sentence 1 - predicted:
\\\textit{vaha}, \textit{bhajan}, \textit{sangraha}, \textit{komal}, \textit{paaswala}, \textit{aytiem}, 10, \textit{par}, \textit{taims}, \textit{sundar (with bindu)}, \textit{hain}, \textit{skul}, \textit{se} 
\vspace{3pt}
\\Sentence 2 - original:
\\\textit{vo}, \textit{bhajansangraha}, \textit{ke} , \textit{paas}, \textit{walaa}, A.T.M., \textit{das}, times, \textit{sundar(with half na)}, \textit{hai}, \textit{skool}, \textit{se}
 \vspace{3pt}
\\A) NO-PENALTY errors:
 \vspace{3pt}
\begin{compactenum}
    \item Colloquial variations:
\\\textit{vaha} vs. \textit{vo}: Colloquial difference - NO PENALTY
\item Compound word handling:
\\\textit{bhajan sangraha} vs. \textit{bhajansangraha}, \textit{paaswala} vs. \textit{paas wala}: Acceptable compound word variations - NO PENALTY
\item Abbreviation variations:
\\\textit{aytiem} vs. A.T.M.: Abbreviation handling - NO PENALTY

\item Numerical variation:
\\10 vs. \textit{das}: Equivalent numerical representation - NO PENALTY
\item Transliterations:
\\\textit{taims} vs times: Acceptable transliteration difference - NO PENALTY
\item Alternate spellings:
\\isundar (with bindu) vs. \textit{sundar (with half na)}: Regional spelling difference - NO PENALTY
\item Transliteration spelling variations:
\\\textit{skul} vs. \textit{skool}: Acceptable transliteration difference - NO PENALTY
\end{compactenum}
 \newpage
\noindent B) MAJOR-PENALTY errors:
 \vspace{3pt}
\begin{compactenum}
    \item \textit{komal} vs. \textit{ke} :
\\Wrong substitution of word – MAJOR PENALTY
\item \textit{par}:
\\Addition of word – MAJOR PENALTY
\end{compactenum}
  \vspace{3pt}
C) MINOR-PENALTY errors:
 \vspace{3pt}
\begin{compactenum}
    \item \textit{hain} vs. \textit{hai}:
\\Small grammatical error that changes singular to plural. – MINOR PENALTY

\end{compactenum}
 \vspace{3pt}
D) EXACT MATCHES:
 \vspace{3pt}
\begin{compactenum}
    \item \textit{se}: Appears identically in both - Not an error
\end{compactenum}
 \vspace{6pt}
 
\noindent \textbf{STRUCTURE OF YOUR RESPONSE:}
 \vspace{6pt}
\\Column 1 – list of no-penalty errors in the format (Word from sentence 1 vs word from
sentence 2 (reason for no penalty), and so on)
\\Column 2 – list of major penalty errors in the format (Word from sentence 1 vs word
from sentence 2 (reason for major penalty), and so on)
\\Column 3 – list of minor penalty errors in the format (Word from sentence 1 vs word
from sentence 2 (reason for minor penalty), and so on)
\\Column 4 – number of no-penalty errors
\\Column 5 – number of major penalty errors
\\Column 6 – number of minor penalty errors

\section{Analysis on English}
\label{d}

 \begin{figure}[htbp] 
  \centering

\includegraphics[width=0.64\linewidth]{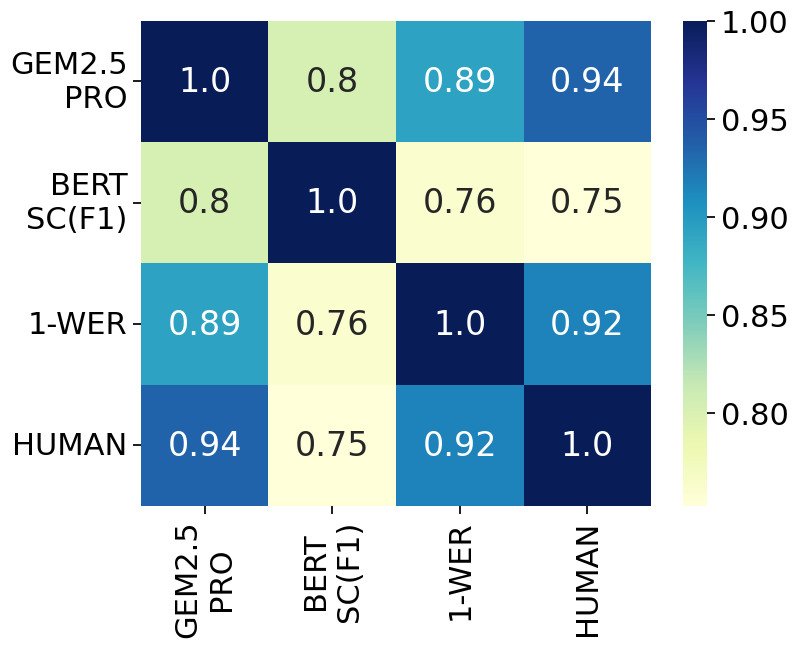}

  \caption{Correlation heatmap for LASER using Gemini 2.5 Pro (Hindi prompt), Human scores, WER and BERTScore(F1) on English data}
  \label{figeng}

  \end{figure}

   \begin{figure*}[t] 
\centering
\includegraphics[clip,width=\linewidth,trim=3.5cm 13cm 4cm 2cm]{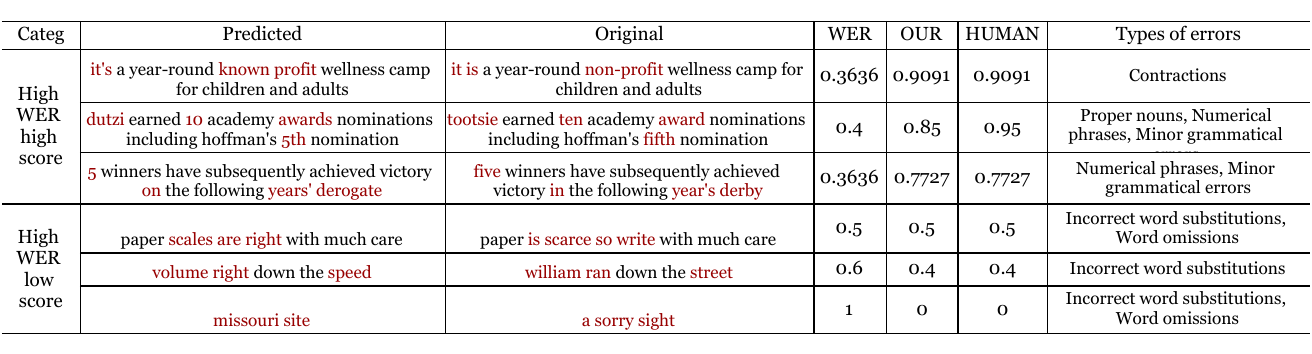}
\vspace{-17pt}
  \caption{Qualitative analysis of high WER English data samples having high and low LASER scores. \textcolor{BrickRed}{Red text} indicates mismatch between the original and predicted transcriptions.}
  \vspace{-10pt}
  \label{figtab2}
  \end{figure*}
  
\noindent \textbf{Single example analysis} 
\vspace{6pt}
\\
We tested LASER (Gemini 2.5 pro) for English data (using the Hindi prompt) with the utterance:
\begin{compactitem}
    \item Original: The colorful bumblebee stung unlucky Priya 3 times on the arm though.
    \item ASR prediction: The colourful bumble-bee strung Pria three times on the arms tho.
\end{compactitem}
\newpage
\noindent LLM response:
\begin{compactitem}
\item Tokens in original: 12
\item 5 No-penalty errors: "colorful vs. colourful" (Alternate spellings), "bumblebee vs. bumble-bee" (compound words), "Priya vs. Pria"(Proper nouns),"3 vs. three" (Numbers),"though vs. tho" (Slang/ Colloquial terms)
\item 1 Minor penalty error: "arm vs. arms" (Small grammatical errors)
\item 2 Major penalty errors: "stung vs. strung" (incorrect word substitution),"omission of unlucky"(omissions/additions)
\item Total penalty: 2.5 and Score: 0.7917
\item Here 1 – WER is 0.3333 (7 substitutions, 1 insertion, WER 0.6667)
\end{compactitem}

\noindent We can observe that the prompt with Hindi examples transfers well to English sentences.
\vspace{6pt}
 
\noindent \textbf{Quantitative analysis}
\vspace{6pt}
 
\noindent Subsequently, we conducted an experiment on 80 samples of the Common-voice English test set, transcribed \cite{commonvoice:2020} using the Whisper \cite{radford2022robustspeechrecognitionlargescale} model. The Hindi prompt was used with the Gemini 2.5 model to get LASER scores for English data. Figure \ref{figeng} depicts the correlations on using the Hindi prompt on English data. It can be observed that LASER scores correlate the best with human evaluations, but the difference between the correlation of WER with human scores and the correlation of LASER with human scores is significantly smaller for English. This indicates that there are fewer no-penalty and minor penalty errors in the English language. This can be due to English being relatively less morphologically complex than Indic languages. Out of the 9 no-penalty error types, only 3 – slang/contractions, proper nouns and numerical phrases - are predominantly observed in English.

\newpage
\noindent \textbf{Qualitative analysis}
 \vspace{6pt}
\\\noindent We  performed a qualitative analysis on English samples. Figure \ref{figtab2} compares samples with high WER (greater than 0.35) and high LASER scores to samples having high WER and low LASER scores. It can be observed that high WER and high LASER score samples contained a high percentage of non-penalizable errors of different types (while retaining the word-pair meanings). In contrast, high WER and low LASER score samples predominantly contained significant semantic word-pair mismatches. Thus, our metric does indeed correct the unfair penalization of semantically identical but syntactically different mismatches between references and ASR predictions. Moreover, human scores are consistent with the LASER scores, thus validating the necessity and utility of our metric.

\label{sec:appendix}

\end{document}